\documentclass[letterpaper, 10 pt, conference]{ieeeconf}  % Comment this line out if you need a4paper

\IEEEoverridecommandlockouts                              % This command is only needed if
\usepackage{graphicx}

\usepackage{subfigure}
\usepackage{multirow}
\usepackage{amssymb}
\usepackage{amsmath}
\usepackage{xcolor}
\usepackage{color}
\usepackage{siunitx}
\usepackage{bbm}
\usepackage[ruled]{algorithm2e}
\usepackage{cite}

\usepackage{prettyref}
\newrefformat{fig}{Figure~\ref{#1}}
\newrefformat{sec}{Section~\ref{#1}}
\newrefformat{tab}{Table~\ref{#1}}
\newrefformat{algo}{Algorithm~\ref{#1}}
\newrefformat{eq}{Equation~\ref{#1}}

\graphicspath{
    {/fig/}
    }
    
\begin{document}

% \title{\LARGE \bf  FF-LOGO: Cross-Modality Point Cloud Registration with Feature Filtering and Local to Global Optimization

% \thanks{
% \textsuperscript{1}Nan Ma, Mohan Wang and Yiheng Han are with the Faculty of Information Technology, Beijing University of Technology, 100 Pingleyuan, Chaoyang District, Beijing 100124, China(e-mail: 
%  \{wangmohan, manan123, hanyiheng\}@emails.bjut.edu.cn).

% \textsuperscript{2}Yong-Jin Liu are with BNRist, MOE-Key Laboratory of Pervasive Computing, Department of Computer Science and Technology, Tsinghua University, Beijing, China.\{liuyongjin@\}tsinghua.edu.cn.
% }
% }
% \author{Nan Ma$^{1}$, Mohan Wang$^{1}$, Yiheng Han$^{\dag}$ and Yong-Jin Liu$^{2}$% <-this % stops a space
% % \thanks{This work was partially supported by ...}% <-this % stops a space
% % \thanks{$^{1}$I.H. Zhan, Y. Han, Y-P Wang and Y-J Liu are with BNRist, MOE-Key Laboratory of Pervasive Computing, Department of Computer Science and Technology, Tsinghua University, Beijing, China.
% %         {\{\tt\small zhanhz20@mails. hyh18@mails. wyp@ and liuyongjin@\}tsinghua.edu.cn}}
% % \thanks{$^{2}$Long Zeng is with the Department of Advanced Manufacturing, Shenzhen International Graduate School, Tsinghua University, Shenzhen, China.
% %         {\tt\small zenglong@sz.tsinghua.edu.cn}}
% % \thanks{$^{*}$Joint first authors  $^{\dag}$Corresponding author}% <-this % stops a space
% }

\title{\LARGE \bf FF-LOGO: Cross-Modality Point Cloud Registration with Feature Filtering and Local to Global Optimization}
\author{Nan Ma$^{1}$, Mohan Wang$^{1}$, Yiheng Han$^{\dag}$ and Yong-Jin Liu$^{2}$% <-this % stops a space
\thanks{\textsuperscript{} This work was supported by the National Natural Science Foundation of China (62371013, 62332019), Beijing Natural Science Foundation (4222025, L222008), QIYUAN LAB Innovation Foundation (Innovation Research) Project (S20210201107), Beijing Municipal Science and Technology (Z221100000222016) and Beijing Hospitals Authority Clinical medicine Development of special funding support (ZLRK202330).}
\thanks{\textsuperscript{1}Nan Ma, Mohan Wang and Yiheng Han are with the Faculty of Information Technology, Beijing University of Technology, 100 Pingleyuan, Chaoyang District, Beijing 100124, China (e-mail: wangmohan@emails.bjut.edu.cn,
 \{manan123, hanyiheng\}@bjut.edu.cn).}
\thanks{\textsuperscript{2}Yong-Jin Liu is with BNRist, MOE-Key Laboratory of Pervasive Computing, Department of Computer Science and Technology, Tsinghua University, Beijing, China. liuyongjin@tsinghua.edu.cn.}
}

\maketitle
\thispagestyle{empty}
\pagestyle{empty}

%%%%%%%%%%%%%%%%%%%%%%%%%%%%%%%%%%%%%%%%%%%%%%%%%%%%%%%%%%%%%%%%%%%%%%%%%%%%%%%%

\begin{abstract}

Cross-modality point cloud registration is confronted with significant challenges due to inherent differences in modalities between sensors. To deal with this problem, we propose FF-LOGO: a cross-modality point cloud registration framework with Feature Filtering and LOcal-Global Optimization. The cross-modality feature correlation filtering module extracts geometric transformation-invariant features from cross-modality point clouds and achieves point selection by feature matching. We also introduce a cross-modality optimization process, including a local adaptive key region aggregation module and a global modality consistency fusion optimization module. Experimental results demonstrate that our two-stage optimization significantly improves the registration accuracy of the feature association and selection module. Our method achieves a substantial increase in recall rate compared to the current state-of-the-art methods on the 3DCSR dataset, improving from 40.59\% to 75.74\%. Our code will be available at https://github.com/wangmohan17/FFLOGO.

\end{abstract}

\section{INTRODUCTION}

Point cloud registration, the task of finding rigid transformations to align two input point clouds, is a pivotal technique in robotics and computer vision \cite{Li_Yang_Lai_Guo_2019}. It finds vital applications in domains such as autonomous driving \cite{Lu_Wan_Zhou_Fu_Yuan_Song_2019}, augmented or virtual reality systems\cite{Ferrari_Cattari_Fontana_Cutolo_2022}, Simultaneous Localization and Mapping (SLAM)\cite{Cattaneo_Vaghi_Valada_2022}, and robotics\cite{Pomerleau_Colas_Siegwart_2015}. Most prior studies have focused on point cloud data of the same modality obtained from identical sensor types, while the potential value of exploring other modalities is not extensively covered. With the advancement of 3D data acquisition technologies, the cost of obtaining point clouds has been reduced, and the methods of acquisition have diversified, including Kinect depth cameras, LiDAR, or Multi-View Stereo (MVS).

Each existing sensor has its specific advantages and limitations when capturing 3D scenes. For instance, LiDAR gauges the distance to obstacles using laser pulses to generate point clouds. Using the high energy of laser pulses, LiDAR generates high-precision point clouds over extended ranges. However, the density of these point clouds is often sparse. Depth cameras estimate depth using infrared or stereo vision techniques, which can produce dense point clouds, but usually within a limited range and with moderate precision. Cross-modality point cloud fusion or registration combines the strengths of multiple sensors, overcoming the limitations of a single sensor, this ensures point cloud acquisition is accurate, efficient, and detailed, resulting in higher-quality visualizations, comparisons, or localizations.

Homogeneous point cloud registration methods face numerous challenges when dealing with cross-modality data. First, due to the diverse mechanisms through which different sensors generate point clouds, the point density distribution varies across modalities. Second, in cross-modality situations, sensor accuracies differ, and outliers can arise from both perceived objects and sensor noise. Additionally, point clouds captured from different kinds of sensors seldom guarantee entirely identical poses and fields of view, making the problem of partial overlap more pronounced in cross-modality point clouds than in homogeneous ones. Existing methods, especially traditional optimization-based ones, can achieve precise results in scenarios with low noise and outliers and are computationally efficient. However, in cross-modality registration, the presence of numerous inaccurate point-to-point correspondences makes finding the optimal solution challenging. Deep learning-based methods, which leverage deep neural networks to extract point cloud features and either base the transformation estimation on these feature correspondences or regress directly from the features to the transformation matrix, offer some robustness. Still, noise and density variations in cross-modality registration can impact feature extraction, resulting in transformation estimation that is often less than satisfactory.

To address these challenges, in this paper, we combine the robustness of deep learning-based cross-modality point cloud feature extraction with the fine-tuning precision advantages of traditional optimization algorithms. We introduce a cross-modality point cloud registration framework with feature filtering and local-global optimization: FF-LOGO. Our method extracts geometric features from cross-modality point clouds with transformation invariance and filters the high feature-coupled uniform point sets and initial pose estimation through a cross-modality feature association filtering module. Subsequent local adaptability key area aggregation modules and global modality-consistent fusion optimization modules perform local-global joint optimization. We observed that the local-global joint optimization process can significantly enhance the pose estimation accuracy of the cross-modality feature association filtering module.

Our main contribution is threefold:
\begin{itemize}
    \item We have devised a cross-modality point cloud registration framework anchored in feature 
correlation filtering module and local-global optimization module to tackle the challenging task of cross-modality point cloud registration.
    
    \item We introduced a novel local-to-global optimization method for cross-modality registration, significantly enhancing the preliminary accuracy derived from feature filtering.
    
    \item We fully leverage the advantages of deep learning in fuzzy correspondence and traditional optimization in pose fine-tuning for cross-modality registration and achieve the state-of-the-art with an improvement from 40.59\% to 75.74\%.
    
\end{itemize}

\section{RELATED WORKS}

\subsection{Conventional Optimization Methods}
The Iterative Closest Point (ICP) \cite{besl1992method} is a classical method in point cloud registration. Its core principle involves iteratively finding the optimal rigid transformation that minimizes the point-to-point distances between two point clouds. While ICP is computationally efficient, dependency on initial pose estimates and vulnerability to outliers pose challenges in cross-modality registration. Variants of the ICP algorithm attempt to enhance its performance from multiple perspectives. For example, trimmed iterative closest point (TriCP) \cite{Chetverikov_Stepanov_Krsek_2005} introduces a trimming rate to effectively select corresponding portions from two datasets for ICP registration, addressing the issue of partial overlaps. Cluster iterative closest point (CICP) \cite{tazir2018cicp}  aims to register point clouds of differing densities produced by varied sensors by matching local surface representations in both source and target point clouds. Building on 4-points congruent sets registration (4PCS) \cite{aiger20084}, Super4PCS \cite{Mellado_Aiger_Mitra_2014} considerably reduces computational demands, offering an efficient solution for global point cloud registration. However, despite these advancements, achieving ideal results in cross-modality point cloud registration remains elusive.

Probabilistic point cloud registration methodologies offer an alternative research direction, seeking to model the deterministic correspondences found in ICP probabilistically. These methods typically employ Gaussian Mixture Models (GMM) to characterize the distribution of point clouds, recasting point cloud registration as an optimization problem of probability density functions. GMM \cite{jian2010robust}stands out as a commonly used approach, exhibiting robustness against significant noise and outliers, albeit with relatively higher computational complexity. FilterReg \cite{Gao_Tedrake_2019}, on the other hand, transforms the correspondence problem in point set registration into a filtering problem using Gaussian filtering, thereby achieving efficient registration. 

Beyond strategies based on GMM, several studies have approached the registration challenge by treating it as a graph-matching problem to address cross-modality discrepancies, CSGM\cite{Huang_Zhang_Fan_Wu_Yuan_2017} as an example, introduces a cross-source point cloud registration approach that preserves both macro and micro-structures. However, its limitation lies in the necessity to segment the point cloud and rely solely on pairwise point matching for graph node alignment. To address these constraints, geometric constraint tensor-based registration (GCTR)\cite{Huang_Fan_Wu_Zhang_Yuan_2019} introduced a cross-modality point cloud registration technique that considers an extended set of neighboring constraints, reformulating point cloud registration into a higher-order graph matching problem. Owing to its inclusion of more constraints, this method exhibits more robust performance in experiments compared to previous graph-matching approaches. Nevertheless, the segmentation process of this approach adds additional computational time, and its performance is intricately tied to hyperparameters.

Recently, a cross-source point cloud registration method based on Geometry Consistent Clustering (GCC) \cite{zhao2023accurate} has reformulated the point cloud registration problem as a consistent clustering process, utilizing adaptive fuzzy clustering to extract structural similarities in point clouds, thus enhancing robustness against outliers. GCC achieves the highest recall rate on the 3DCSR\cite{huang2021comprehensive} dataset at 40.59\%, but remains less than satisfactory. Our proposed method, without compromising efficiency, elevates the recall accuracy to 75.74\%.

\subsection{Deep neural network methods}

The achievements of deep neural networks in three-dimensional geometry, such as PointNet \cite{Charles_Su_Kaichun_Guibas_2017} and DGCNN \cite{Wang_Sun_Liu_Sarma_Bronstein_Solomon_2019}, have propelled advancements in deep point cloud registration. At the heart of these methods lies the concept of leveraging deep neural networks to extract features from cross-source point clouds, either basing registrations on these feature correspondences or directly regressing transformation matrices from the features themselves. Feature learning techniques, SpinNet \cite{Ao_Hu_Yang_Markham_Guo_2021} as an example aims to extract robust point descriptors through tailored neural network designs. However, its reliance on a voxelization preprocessing step poses challenges in the context of cross-modality point clouds. Another method such as D3Feat \cite{Bai_Luo_Zhou_Fu_Quan_Tai_2020} necessitates the construction of features based on k-nearest neighbors, but this descriptor tends to falter when faced with significant density disparities. Beyond these point descriptor-focused methodologies, several strategies emphasize feature matching. Deep Global Registration (DGR) \cite{Choy_Dong_Koltun_2020} devises a UNet architecture for discerning whether a point pair corresponds. This process reinterprets the feature-matching dilemma as a binary classification task. Transformation learning approaches, as an alternative line of investigation, directly estimate transformations via neural networks. Feature-metric registration (FMR)\cite{Huang_Mei_Zhang_2020} introduces a feature metric registration technique, aligning two point clouds by minimizing their feature metric projection error. However, due to the significant disparities in cross-modality point clouds, registration methods based solely on neural networks are susceptible to outliers and struggle to effectively extract structural consistency features of cross-modality point clouds, leading to suboptimal registration performance. Our method builds upon feature extraction to perform stepwise pose optimization at both local and global levels, ensuring the stability of the algorithm and the accuracy of the registration results.

 \begin{figure*}[t]
  \centering
  {\includegraphics[width=1.8\columnwidth]{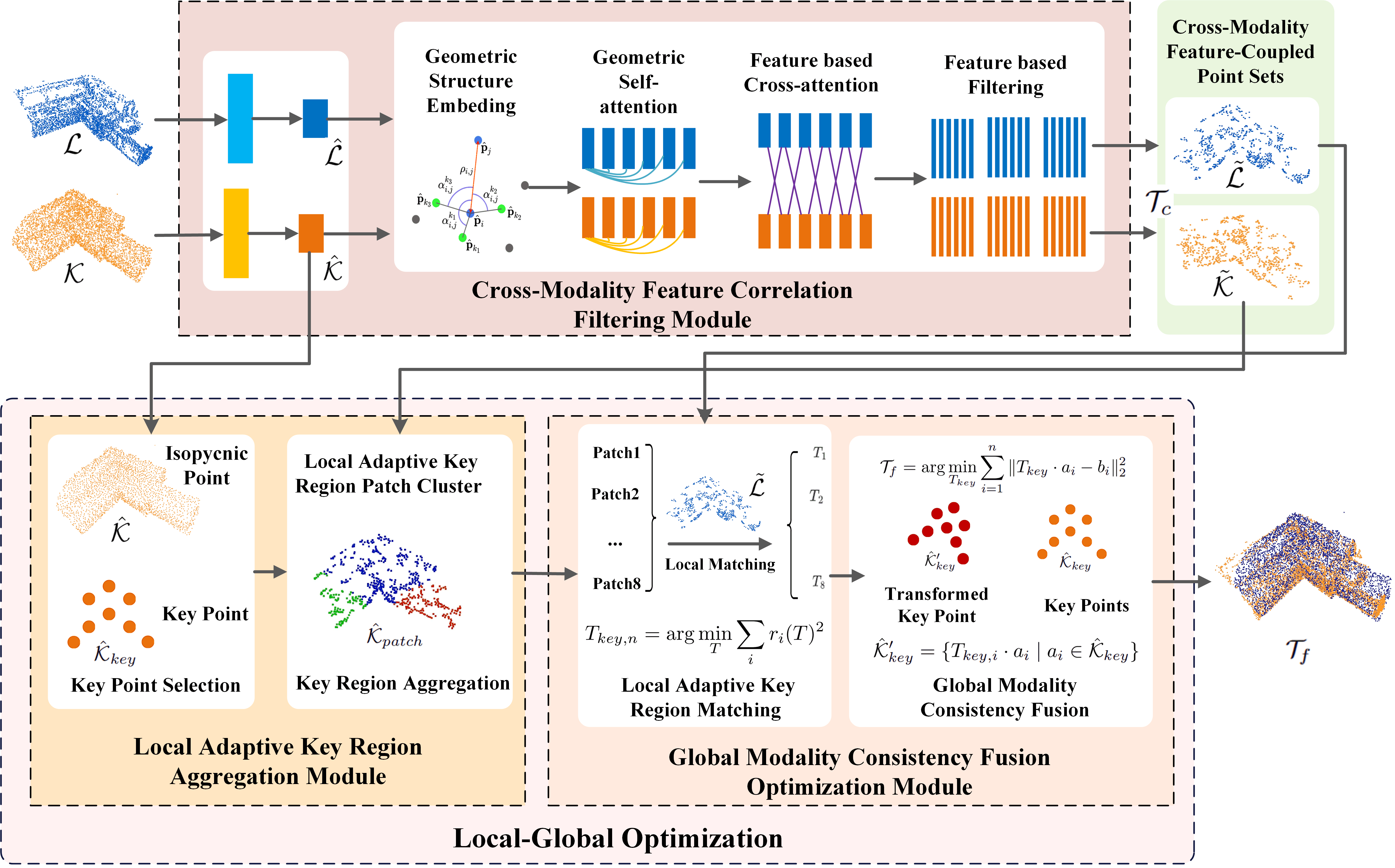}}
  \vspace{-10px}
  \caption{\textbf{Overview of the proposed pipeline. The Cross-Modality Feature Correlation Filtering Module extracts and filters feature-correlated points, obtaining an initial pose estimation. Key regions identified by The Local Adaptive Key Region Aggregation Module are then optimized with the Global Modality Consistency Fusion Optimization Module to achieve the final optimized registration.}}
  \label{fig:pipline}
  \vspace{-20px}
  \end{figure*}

\section{METHODOLOGY}

The comprehensive architecture of our proposed method is illustrated in Figure~\ref{fig:pipline}, encompassing three integral modules. The Cross-Modality Feature Correlation Filtering Module operates on the original point clouds, \( \mathcal{K} \) and \( \mathcal{L} \), which are from different modalities. Voxel downsampling is performed to mitigate density discrepancies, preventing adverse effects on subsequent feature extraction, it produces isopycnic points, meaning the points are in equivalent densities, denoted as \( \hat{\mathcal{K}} \) and \( \hat{\mathcal{L}} \). Building upon this foundation, the module isolates points with heightened feature confidence and gets the cross-modality feature-coupled point sets, as \( \tilde{\mathcal{K}} \) and \( \tilde{\mathcal{L}} \). Simultaneously, a coarse-grained pose transformation estimation \( \mathcal{T}_c \) is derived via feature matching techniques. The Local Adaptive Key Region Aggregation Module identifies prominent key points, \( \hat{\mathcal{K}}_{key} \), from \( \hat{\mathcal{L}} \), extracting a subset of the point cloud, \( \hat{\mathcal{K}}_{patch} \), which encapsulates key local regions with a pronounced adaptive nature. The Global Modality Consistency Fusion Optimization Module fuses the point set \( \tilde{\mathcal{L}} \), discerned from the initial module, with the adaptive key local region sub-point cloud \( \hat{\mathcal{K}}_{patch} \) from the secondary module. By deploying a modality-consistent optimization strategy that transitions from local nuances to a global perspective, the coarse-grained pose transformation estimate \( \mathcal{T}_{c} \) undergoes meticulous refinement, yielding the definitive registration outcome \( \mathcal{T}_{f} \). The details of each module will be provided in subsections.

\subsection{The Cross-Modality Feature Correlation Filtering Module}

Our Cross-Modality Feature Correlation Filtering Module is mainly inspired by the GeoTransformer \cite{qin2022geometric}. Operating under identical physical environments, the geometric position encoding self-attention module of GeoTransformer effectively extracts geometry transformation-invariant features from point clouds. This capability robustly handles challenges in cross-modality point cloud matching, such as outliers, anomalies, and partial overlaps. To elaborate on the methodology, the module encompasses three distinct components: the construction of Geometric Self-attention, the extraction of features through the Geometric Transformer, and the subsequent process of feature matching and selection.

Initially, we use Geometric Self-attention to learn the feature and global correlation in the geometric space between isopycnic points. The geometric structure embedding, at its core, aims to utilize consistent angular and distance relationships present in cross-modality point clouds from the same scene. For point set (\( \hat{\mathcal{K}} \) or \( \hat{\mathcal{L}} \)), the geometric structure embedding consists of pair-wise distance embedding and triplet-wise angular embedding:

\begin{equation}
     \mathbf{e}_{i, j} = \mathbf{e}_{i, j}^D \mathbf{W}^D + \max_x \left\{ \mathbf{e}_{i, j, x}^A \mathbf{W}^A \right\} 
\end{equation}

\( \mathbf{W}^D \) and \( \mathbf{W}^A \) are projection matrices for distance and angular embeddings respectively, \( \mathbf{e}_{i, j} \) is the geometric structure embedding, \( \mathbf{e}_{i, j}^D \) is the pair-wise distance embedding, and \( \mathbf{e}_{i, j}^A \) is the triplet-wise angular embedding.

The pair-wise distance embedding, \( \mathbf{e}_{i, j}^D \), is calculated as:

\begin{equation}
\begin{aligned} e_{i, j, 2k}^D & = \sin \left( \frac{d_{i, j} / \sigma_d}{10000^{2k/d_t}} \right) \\ e_{i, j, 2k+1}^D & = \cos \left( \frac{d_{i, j} / \sigma_d}{10000^{2k/d_t}} \right) \end{aligned}
\end{equation}

\( d_{i, j} \) represents the Euclidean distance between points \( \hat{\mathbf{p}}_i \) and \( \hat{\mathbf{p}}_j \), \( \sigma_d \) is a hyperparameter to adjust distance variations, and \( d_t \) is the dimensionality of the data. The triplet-wise angular embedding, \( \mathbf{e}_{i, j}^A \), is given by:

\begin{equation}
\begin{aligned} e_{i, j, x, 2l}^A & = \sin \left( \frac{\alpha_{i, j}^x / \sigma_a}{10000^{2l/d_t}} \right) \\ e_{i, j, x, 2l+1}^A & = \cos \left( \frac{\alpha_{i, j}^x / \sigma_a}{10000^{2l/d_t}} \right) \end{aligned}
\end{equation}

\( k \) neighboring points are initially chosen for \( \hat{\mathbf{p}}_i \), forming the point set \( \mathcal{X}_i \). For \( \hat{\mathbf{p}}_x \) within \( \mathcal{X}_i \), \( \alpha_{i, j} \) is calculated as: \( \alpha_{i, j}^x = \angle (\Delta_{x, i}, \Delta_{j, i}) \), with \( \Delta_{i, j} \) defined as \( \hat{\mathbf{p}}_i - \hat{\mathbf{p}}_j \).

In the subsequent process, the Geotransformer\cite{qin2022geometric} network is utilized to compute self-attention and cross-attention based on geometric structure embedding. This yields features \( \hat{\mathcal{H}}^\mathcal{K} \) and \( \hat{\mathcal{H}}^\mathcal{L} \) for the point sets \( \hat{\mathcal{K}} \) and \( \hat{\mathcal{L}} \), respectively.

The final step involves point cloud selection based on their features. First, \( \hat{\mathcal{H}}^\mathcal{K} \) and \( \hat{\mathcal{H}}^\mathcal{L} \) are normalized onto a unit hypersphere. The Gaussian correlation matrix \( S \) is then computed, with entries \( s_{i,j} \in S \) defined as:

\begin{equation}
s_{i, j} = \exp \left( -\left\| \hat{\mathbf{h}}_i^{\mathcal{K}} - \hat{\mathbf{h}}_j^{\mathcal{L}} \right\|_2^2 \right)
\end{equation}

% equation transforms the square of the Euclidean distance between \( \hat{h}_i^K \) and \( \hat{h}_j^L \) into a similarity metric, representing the similarity of corresponding relationships between isopycnic point sets \( \hat{\mathcal{K}} \) and \( \hat{\mathcal{L}} \). From the Gaussian correlation matrix \( S \), the top-k entries are chosen as overlapping point correspondences \( \hat{\mathcal{C}} \):

%\begin{equation}
%\hat{\mathcal{C}} = \left\{ \left( \hat{\mathbf{p}}_{x_i}, \hat{\mathbf{q}}_{y_i} \right) \mid \left( x_i, y_i \right) \in \operatorname{topk}_{x, y} \left( \bar{s}_{x, y} \right) \right\}
%\end{equation}

%Using the overlapping point correspondences, the transformation \( \mathcal{T}_c = \left\{ \mathbf{R}_i, \mathbf{t}_i \right\} \) is derived:

%\begin{equation}
 %\mathbf{R}_i, \mathbf{t}_i = \min_{\mathbf{R}, \mathbf{t}} \sum_{\left( \tilde{\mathbf{p}}_{x_j}, \tilde{\mathbf{q}}_{y_j} \right) \in \mathcal{C}_i} \left\| \mathbf{R} \cdot \tilde{\mathbf{p}}_{x_j} + \mathbf{t} - \tilde{\mathbf{q}}_{y_j} \right\|_2^2 
%\end{equation}

%\( \tilde{\mathbf{p}}_{x_j} \) and \( \tilde{\mathbf{q}}_{y_j} \) are the global dense point correspondences in \( \hat{\mathcal{C}} \). The corresponding point pairs in \( \hat{\mathcal{C}} \) are the cross-modality feature-coupled isopycnic point sets \( \tilde{\mathcal{P}} \) and \( \tilde{\mathcal{Q}} \). \( \mathcal{T}_c \) represents the initial transformation derived from feature correlation matching.

% \begin{equation}
    
% \end{equation}

\subsection{The Local Adaptive Key Region Aggregation Module}

The role of the Local Adaptive Key Region Aggregation module is to aggregate several locally representative point cloud patches from the points to prepare for the global modality consistency fusion optimization. The extracted local point cloud patches should be geometric representative, dispersed throughout the point cloud, and contain ample local features.

Initially, from the point set \( \hat{\mathcal{K}} \), we select dispersed and geometrically representative key points within the point cloud. Since the purpose is to capture the maximal geometric feature of the entire point cloud with the minimal number of points, we adopt the approach outlined in PVNET\cite{peng2019pvnet}, utilizing the Farthest Point Sampling (FPS) algorithm to select \( n \) key points. By iteratively selecting points that are maximally distant from those already chosen, FPS ensures uniform coverage across the spatial extent of the point cloud. Assuming our point cloud, denoted as set \( P \), where each point is represented as a vector, and the set of selected points is \( S \). To initialize the set of key points, we calculate the geometric centroid of the point cloud. For every remaining point \( p \in P \), we compute its minimum distance to all points in set \( S \):
\begin{equation}
d(p, S) = \min_{s \in S} \| p - s \|_2
\end{equation}

We then select the point with the maximum distance:
\begin{equation}
p_{\text{next}} = \arg \max_{p \in P \backslash S} d(p, S)
\end{equation}

The chosen point \( p_{\text{next}} \) is added to the set \( S \) and removed from set \( P \), continuing this process until the size of the set reaches \( n \). This results in the key point set \( \hat{\mathcal{K}}_{key} \). For accuracy and efficiency, we adopt \( n = 8 \). After extracting the key points, we employ the KNN algorithm to aggregate neighboring points around the key points from \( \hat{\mathcal{K}} \) to form the Local Adaptive Key Region point set \( \hat{\mathcal{K}}_{patch} \), which supplements the local information of the point cloud.

\subsection{The Global Modality Consistency Fusion Optimization Module}

Due to discrepancies between training and validation sets, as well as losses in local information arising from feature extraction, registration results from feature-matching methods frequently yield reduced accuracy or even outright failures. It is crucial to impose additional constraints on the registration results through optimization techniques following cross-modality feature alignment to enhance accuracy and stability. In consideration of this, we have developed an optimization method that transitions from local adaptive key region matching to global modality consistency fusion.

The point set \( \hat{\mathcal{L}} \), processed through the Cross-Modality Feature Correlation Filtering Module, yields the cross-modality feature-coupled point set \( \tilde{\mathcal{L}} \), representing the point cloud post-feature selection. The point set \( \hat{\mathcal{K}} \), processed through the Local Adaptive Key Region Aggregation Module, gives the local adaptive key region point set \( \hat{\mathcal{K}}_{patch} \), symbolizing the key point cloud region containing representative local information. The \( \tilde{\mathcal{L}} \) is then matched with each local adaptive key region in \( \hat{\mathcal{K}}_{patch} \) to compute the point-to-plane residuals and iteratively map to obtain the optimal transformation pose. Specifically, for every point \( a_i \) in the \( n \) th set of \( \hat{\mathcal{K}}_{patch} \), locate the corresponding plane \( j \) in the point cloud \( \tilde{\mathcal{L}} \) with the shortest distance, where \( \vec{n}_j \) is the normal vector of plane \( j \):

\begin{equation}
j^*(a_i) = \arg \min_j |(a_i - b_{j1}) \cdot \vec{n}_j |
\end{equation}

For each point \( a_i \) and its corresponding plane \( j \), compute the point-to-plane distance residual. Where \( T \) is the initial pose estimation obtained from The Cross-Modality Feature Correlation Filtering Module, and \( T(a_i) \) represents the position of point \( a_i \) after transformation \( T \):

\begin{equation}
r_i = (T(a_i) - b_{j1}) \cdot \vec{n}_j
\end{equation}

Minimize the distance residuals \( r_i \) for all points to their corresponding planes by solving the least squares problem:
\begin{equation}
T_{key,n} = \arg \min_{T} \sum_i r_i(T)^2
\end{equation}

This provides the optimal modality-consistent transformation \( T_{key,n} \) between the \( n \) th \( \hat{\mathcal{K}}_{patch} \) and the cross-modality feature-coupled point set \( \tilde{\mathcal{L}} \).

After obtaining the local optimal transformations, a global key point least squares optimization is necessary to integrate local modality-consistent adjustments. We transform the corresponding key points from the local adaptive key region point set according to the local transformation estimate \( T_{key} \) to get the transformed point set \( \hat{\mathcal{K}}_{key}' \):
\begin{equation}
\hat{\mathcal{K}}_{key}' = \{ T_{key,i} \cdot a_i \mid a_i \in \hat{\mathcal{K}}_{key} \}
\end{equation}

For each corresponding point pair in \( \hat{\mathcal{K}}_{key} \) and \( \hat{\mathcal{K}}_{key}' \), we solve the following least squares problem to minimize the squared distance error for all points:
\begin{equation}
\mathcal{T}_f = \arg \min_{T_{key}} \sum_{i=1}^n \| T_{key} \cdot a_i - b_i \|_2^2
\end{equation}

This results in the final optimized transformation \( \mathcal{T}_f \) through the local-to-global modality-consistent transformation estimation. We summarize the proposed method of The Global Modality Consistency Fusion Optimization in Algorithm~\ref{alg:LOGO}.

\vspace{-10px}
\begin{algorithm}
\caption{The Global Modality Consistency Fusion Optimization}
\label{alg:LOGO}
\LinesNumbered
\SetKwInOut{Input}{Input}
\SetKwInOut{Output}{Output}
\SetKwInOut{Initialize}{Initialize}

\Input{Isopycnic point sets \( \hat{\mathcal{K}} \), \( \hat{\mathcal{L}} \)}
\Output{Optimal transformation \( \mathcal{T}_{f} \)}
\Initialize{Transformed point set \( \hat{\mathcal{K}}_{key}^{\prime} \), Point-to-plane residual \( r_i \), Transformation estimates \( T_{key,n} \)}

\BlankLine
\( \tilde{\mathcal{L}} \leftarrow \) Process \( \hat{\mathcal{L}} \) through The Cross-Modality Feature Correlation Filtering Module\;
\( \hat{\mathcal{K}}_{patch} \leftarrow \) Process \( \hat{\mathcal{K}} \) through The Local Adaptive Key Region Aggregation Module\;

\ForEach{point set \( \hat{\mathcal{K}}_{patch,n} \) in \( \hat{\mathcal{K}}_{patch} \)}{
    \ForEach{point \( a_i \) in \( \hat{\mathcal{K}}_{patch,n} \)}{
        Find plane \( b_{j1}, b_{j2}, b_{j3} \) in \( \tilde{\mathcal{L}} \) with minimum distance to \( a_i \)\;
        Compute the point-to-plane distance residual \( r_i \)\;
    }
    Solve for \( T_{key,n} \) to minimize all \( r_i \)\;
}

Transform \( \hat{\mathcal{K}}_{key} \) using \( T_{key} \) to get \( \hat{\mathcal{K}}_{key}^{\prime} \)\;
\ForEach{point pair in \( \hat{\mathcal{K}}_{key} \) and \( \hat{\mathcal{K}}_{key}^{\prime} \)}{
    Solve least squares problem to minimize squared distance error\;
}
Compute the final transformation \( \mathcal{T}_{f} \) using local-to-global modality-consistent estimation\;

\end{algorithm}
\vspace{-10px}

\section{EXPERIMENT}

\subsection{Implementation Details}

\subsubsection{Parameters}
In the cross-modality feature correlation selection module, the voxel size for voxel-based subsampling during dense point selection is set to \(0.05m\).  We utilized the Adam optimizer \cite{Kingma_Ba_2014} to train our network over a span of 40 epochs, employing the 3DMatch\cite{Zeng_Song_Niessner_Fisher_Xiao_Funkhouser_2017} dataset. The training parameters included a batch size of 1, a weight decay set at $10^{-6}$, and an initial learning rate of $10^{-4}$, which undergoes an exponential decay at a rate of 0.05 with every passing epoch. Regarding the mutual top-k selection in the point correspondences filtering, we set the hyper-parameter $k$ :  $k$ =1 for 250, 500, and 1000 matches, $k$ =2 for 2500 matches, and $k$ = 3 for 2500 matches. This configuration helps in regulating the number of point correspondences. In the local adaptability key region aggregation module, the number of key points selected for the set \(\hat{\mathcal{K}}_{key}\) is \( n = 8 \). The aggregation radius for gathering neighboring points around the key points from the set \(\hat{\mathcal{K}}\) is \( d(p, S)_{max} \times 1.5 \). All experiments are conducted on a machine equipped with a single RTX 4090 graphics card and an Intel Core i5-13600KF CPU.

\subsubsection{Dataset}
The proposed algorithm and baseline methods are evaluated on the 3DCSR dataset \cite{huang2021comprehensive}. Point clouds in this dataset originate from three distinct modalities: LiDAR, Kinect, and camera sensors. Point clouds produced by the LiDAR equipment are relatively sparse, while the Kinect point clouds, generated by the Kinect depth camera, are dense and uniform. The third modality data is constructed from a series of indoor 2D RGB images using the Structure from Motion (SfM) approach. This dataset provides ground truth transformations for aligning either LiDAR or SfM geometry with dense Kinect geometry. It encompasses the most common objects or scenes found in indoor working environments. In total, the dataset comprises 202 point cloud pairs, with 37 scenes captured by Kinect and RGB cameras, and 165 scenes acquired by LiDAR and Kinect sensors.

\subsection{Evaluation Metrics}

For rigorous evaluation, we adopt two key metrics to quantify the quality of registration:

\[
\begin{aligned} 
\operatorname{RE}(\hat{\boldsymbol{R}}, \boldsymbol{R}) & =\frac{180}{\pi} \arccos \left(\frac{1}{2}\left\{\operatorname{Tr}\left(\hat{\boldsymbol{R}} \boldsymbol{R}^T\right)-1\right\}\right), \\ 
\mathrm{TE}(\hat{\boldsymbol{t}}, \boldsymbol{t}) & =\|\hat{\boldsymbol{t}}-\boldsymbol{t}\|_2,
\end{aligned}
\]

Where Rotation Error (RE) represents the geodesic distance within \(SO(3)\) and Translation Error (TE) signifies the Euclidean distance in \(R^3\). These metrics effectively evaluate the discrepancies in rotation and translation between the estimated outcome \((\hat{R}, \hat{t})\) and the established ground truth \((R, t)\).

For all methods, we conducted 10 independent experiments to assess their average performance and recall rates. The registration recall rate is calculated as the ratio of the number of point cloud pairs with RE less than 15$^{\circ}$ and a TE less than 0.3m to the total number of pairs. Consistent with some literature, such as GCC \cite{zhao2023accurate}, when evaluating average performance, we only consider point cloud pairs that were successfully recalled. This is because pairs that fail to be recalled deviate significantly from the baseline data, rendering their performance metrics potentially unreliable.

\subsection{Performance}

The quantitative results are shown in Table~\ref{tab:method_comparison}. In addition to the latest cross-source registration methods, performance tests of many other homomodal point cloud registration benchmarks are also reported. Our method significantly outperforms the current state-of-the-art method, GCC, in recall rate, increasing from 40.59\% to 75.74\%. TE and RE are slightly higher than the current lowest method because they are calculated only based on successfully recalled samples.

Conventional optimization methods such as RANSAC\cite{fischler1981random}, ICP\cite{besl1992method}, and its variants like TrICP\cite{Chetverikov_Stepanov_Krsek_2005}, CICP\cite{tazir2018cicp}, PICP\cite{jubran2021provably}, and Super4PCS\cite{Mellado_Aiger_Mitra_2014}, do not show significant advantages in recall rate and estimation error. The graph matching-based GCTR\cite{Huang_Fan_Wu_Zhang_Yuan_2019} has an even lower recall rate, while methods based on Gaussian Mixture Model (GMM), such as FilterReg\cite{Gao_Tedrake_2019}, demonstrate relatively higher recall rates and moderate estimation errors. Apart from IDAM\cite{li2020iterative}, deep neural network-based methods such as DGR\cite{Choy_Dong_Koltun_2020}, and FMR\cite{Huang_Mei_Zhang_2020} all exhibit higher-than-average recall rates and relatively lower estimation errors compared to conventional optimization methods. Notably, the avant-garde cross-modality point cloud registration technique, GCC\cite{zhao2023accurate}, outperforms the DGR method, which harnesses deep neural network feature mapping in recall metrics, setting a new benchmark. Nonetheless, its sub-50\% recall rate underscores the inherent challenges plaguing the domain of cross-modality point cloud registration. Existing methods still face significant hurdles in addressing cross-modality point cloud registration generalization. Our method offers a substantial improvement in generalization, achieving a recall rate of 75.74\%. Furthermore, while maintaining a high recall rate, our method also retains a relatively low level of estimation error compared to current methods, indicating its advantages in both generalization and accuracy.
\vspace{-10px}
\begin{table}[h]
    \centering
    \caption{Quantitative comparisons on the 3DCSR dataset.}
    \vspace{-5px}
    \begin{tabular}{l|c|c|c}
    \hline
    Method    & Recall(\%)    & TE(m)       & RE($^{\circ}$)       \\
    \hline
    ICP\cite{besl1992method}       & 24.3      & 0.38     & 5.71     \\
    RANSAC\cite{fischler1981random}    & 3.47      & 0.13     & 8.30     \\
    TrICP\cite{Chetverikov_Stepanov_Krsek_2005}     & 7.92      & 0.18     & 6.40     \\
    CICP\cite{tazir2018cicp}      & 2.48      & 0.28     & 8.28     \\
    PICP\cite{jubran2021provably}      & 4.45      & 0.29     & 10.85    \\
    Super4PCS\cite{Mellado_Aiger_Mitra_2014} & 6.93      & 0.24     & 6.38     \\
    GCTR\cite{Huang_Fan_Wu_Zhang_Yuan_2019}      & 0.50      & 0.17     & 7.46     \\
    GMM\cite{jian2010robust}       & 9.41      & 0.18     & 7.92     \\
    FilterReg\cite{Gao_Tedrake_2019} & 30.96     & 0.10     & 2.45     \\
    DGR\cite{Choy_Dong_Koltun_2020}       & 36.60     & \textbf{0.04} & 4.26     \\
    IDAM\cite{li2020iterative}      & 1.98      & 0.13     & 11.37    \\
    FMR\cite{Huang_Mei_Zhang_2020}       & 17.80     & 0.10     & 4.66     \\
    GCC\cite{zhao2023accurate}       & 40.59     & 0.06     & \textbf{2.21} \\
    FF-LOGO (Ours)      & \textbf{75.74} & 0.09     & 2.85     \\
    \hline
    \end{tabular}
    \label{tab:method_comparison}
    \vspace{-15px}
\end{table}

\subsection{Ablations}

To analyze the efficacy of the proposed local-global optimization process, we present an ablation study in Table~\ref{tab:ablation}. Specifically, Feature Filtering (FF) represents the cross-modality feature association filtering module, Local-Global Optimization (LOGO) embodies the combined optimization workflow of the local adaptability key region aggregation module and the global modality consistency fusion optimization module. At the same time, Global Optimization (GO) denotes a module that encompasses solely global optimization, serving as a replacement for LOGO. Employing the GO module independently increased the recall rate from 61.88\% to 65.34\%, whereas utilizing the LOGO module, in comparison to using the GO module alone, further raised it from 65.34\% to 75.74\%. The results demonstrate that the local-global optimization method of LOGO significantly improves the initial registration accuracy of the FF module compared to the GO module. Besides, in comparison to all existing approaches, FF with deep learning methods exhibits significant potential for enhancing the processing capability of cross-modality point clouds. However, it remains a coarse registration method concerning the specific challenge of cross-modality point cloud alignment and experiments show that our Local-Global optimization method can accomplish fine registration on this basis and yield further enhanced performance improvements.  These findings validate the effectiveness of the local-global optimization procedure in the context of cross-modality point cloud registration.

\vspace{-10px}
\begin{table}[h]
    \centering
    \caption{Ablation study on FF-LOGO}
    \vspace{-5px}
    \begin{tabular}{c|c|c|c}
    \hline
    FF & GO & LOGO  & Recall(\%) \\
    \hline
    \checkmark & &  & 61.88 \\
    \checkmark & \checkmark &  & 65.34 \\
    \checkmark & & \checkmark  & 75.74 \\
    \hline
    \end{tabular}
    \label{tab:ablation}
    \vspace{-15px}
\end{table}

\subsection{Application}
 \begin{figure}[t]
  \centering
  {\includegraphics[width=1.0\columnwidth]{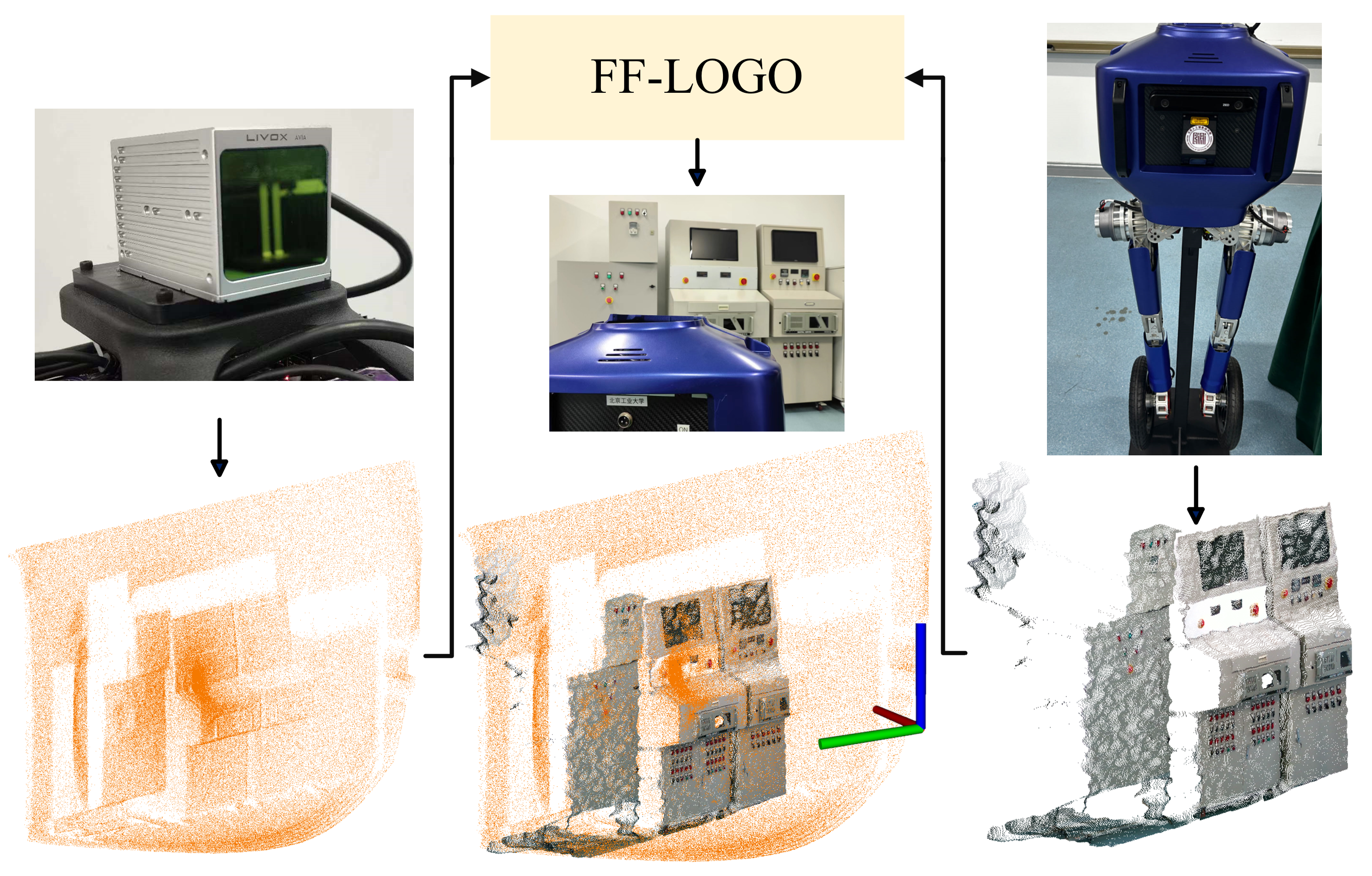}}
  \vspace{-15px}
  \caption{\textbf{Application of FF-LOGO}}
  \label{fig:app}
  \vspace{-20px}
  \end{figure}

To demonstrate the practical application value of FF-LOGO in cross-modality point cloud registration, we deployed the algorithm on a bipedal wheeled robot for cross-modality localization tests. As illustrated in Figure~\ref{fig:app}, we initially constructed a high-precision point cloud map using a LIDAR scanner. Subsequently, point clouds generated in real-time from the stereo camera equipped on the wheeled robot were registered with the high-quality LIDAR point clouds to find the accurate pose of our robot. Experimental results validate the robustness and high accuracy of our approach, with very few instances of registration failure and a localization error of less than 10mm. FF-LOGO holds significant potential for many robotic tasks such as localization, point cloud completion, and registration for cross-modality scenarios.

% \vspace{-5px}
\section{CONCLUSION}
% \vspace{-5px}

In this paper, we introduced a novel framework combining feature filtering and local-global optimization, resulting in robust and accurate registration. Our method fully leverages the advantages of deep learning and traditional optimization, achieving a significant improvement in registration accuracy, as evidenced by a substantial increase in the recall rate compared to state-of-the-art methods on the 3DCSR dataset. In the future, we will explore the potential of our method for general point cloud registration.
% In this paper, we proposed a pose estimation pipeline KRF that combines estimation methods, a point cloud completion network and a Color Iterative KeyPoint method. Experiments show that all novel components are effective, and our method outperforms the state-of-the-art methods on YCB-Video and Occlusion LineMOD datasets.

\addtolength{\textheight}{-1.9cm}   % This command serves to balance the column lengths

%%%%%%%%%%%%%%%%%%%%%%%%%%%%%%%%%%%%%%%%%%%%%%%%%%%%%%%%%%%%%%%%%%%%%%%%%%%%%%%%

%%%%%%%%%%%%%%%%%%%%%%%%%%%%%%%%%%%%%%%%%%%%%%%%%%%%%%%%%%%%%%%%%%%%%%%%%%%%%%%%
%\section*{APPENDIX}


%\section*{ACKNOWLEDGMENT}

%The preferred spelling of the word.

\bibliographystyle{IEEEtran}
\bibliography{rootbib}

\end{document}